# Training deep learning based dynamic MR image reconstruction using synthetic fractals

Anirudh Raman[1], Olivier Jaubert[1], Mark Wrobel[1], Tina Yao[1], Ruaraidh Campbell[1], Rebecca Baker[1], Ruta Virsinskaite[2], Daniel Knight[1,2], Michael Quail[3], Jennifer Steeden[1], Vivek Muthurangu[1]

Affiliations:

1. UCL Centre for Translational Cardiovascular Imaging, University College London, 30 Guilford St, London, WC1N 1EH, UK
2. Department of Cardiology, Royal Free London NHS Foundation Trust, London, NW3 2QG, UK
3. Institute of Cardiovascular Science, University College London, London, WC1E 6HX, UK

**Abstract**
*Purpose:*
To investigate whether synthetically generated fractal data can be used to train deep learning (DL) models for dynamic MRI reconstruction, thereby avoiding the privacy, licensing, and availability limitations associated with cardiac MR training datasets.

*Methods:*
A training dataset was generated using quaternion Julia fractals to produce 2D+time images. Multi-coil MRI acquisition was simulated to generate paired fully sampled and radially undersampled k-space data. A 3D UNet deep artefact suppression model was trained using these fractal data (F-DL) and compared with an identical model trained on cardiac MRI data (CMR-DL). Both models were evaluated on prospectively acquired radial real-time cardiac MRI from 10 patients. Reconstructions were compared against compressed sensing(CS) and low-rank deep image prior (LR-DIP). All reconstrctuions were ranked for image quality, while ventricular volumes and ejection fraction were compared with reference breath-hold cine MRI.

*Results:*
There was no significant difference in qualitative ranking between F-DL and CMR-DL (p=0.9), while both outperformed CS and LR-DIP (p<0.001). Ventricular volumes and function derived from F-DL were similar to CMR-DL, showing no significant bias and accptable limits of agreement compared to reference cine imaging. However, LR-DIP had a signifcant bias (p=0.016) and wider lmits of agreement.

*Conclusion:*
DL models trained using synthetic fractal data can reconstruct real-time cardiac MRI with image quality and clinical measurements comparable to models trained on true cardiac MRI data. Fractal training data provide an open, scalable alternative to clinical datasets and may enable development of more generalisable DL reconstruction models for dynamic MRI.

**Introduction**

Real-time MRI enables assessment of dynamic changes without physiological gating but requires significant data undersampling to ensure sufficient spatiotemporal resolution. Compressed Sensing (CS) is widely used to reconstruct artefact-free images from undersampled real-time data[1], but is limited by time-consuming iterative reconstructions. More recently, Deep Learning (DL) reconstruction methods have demonstrated faster inference and superior image quality compared to CS [2].

Fast reconstruction is particularly important in cardiac applications, where operators often need to review the latest scan prior to planning subsequent acquisitions. One DL approach for rapid reconstruction of real-time MRI data is deep artefact suppression, a 'post-processing' method that has shown strong performance when combined with non-Cartesian trajectories. However, this approach is limited by the need for large application-specific MRI training datasets, which can be difficult to obtain. Even where large datasets are available locally, data protection regulations often limit data-sharing. We have previously shown that open-source videos can be used to train DL-based MRI reconstructions [3], but several issues restrict their use, including: (i) Clinical scepticism towards DL reconstructions trained on highly out-of-distribution content (e.g. animals, vehicles, or landscapes); and (ii) Licensing restrictions on video datasets, which may complicate commercial deployment.

This motivates the exploration of alternative, synthetic training data that are structurally rich yet unconstrained by privacy or licensing limitations. Consequently, we propose an alternate source of training data; time-varying Julia set fractals. These data are structurally complex but abstract, making them seem less incongruous than natural videos, while retaining high spatial and temporal complexity and remaining fully open-source.

The aims of this proof-of-concept study are: (i) To demonstrate the feasibility of generating a synthetic fractal dataset; (ii) To use this data to train a deep artefact suppression model for dynamic MRI reconstruction; (iii) To evaluate the approach on prospectively acquired radial real-time cardiac MRI data; and (iv) To compare performance against other reconstruction methods, including deep artefact suppression trained on cardiac MRI data, CS and Deep Image Prior (DIP).

**Methods**

*Fractal Training Data*

Synthetic Julia set MRI training data were generated by: (i) Producing time-varying Julia set fractals in quaternion space; (ii) Converting them to complex-valued dynamic images; and (iii) Simulating multi-coil acquisition to obtain k-space data. An overview is shown in Figure 1 with all code available at: https://github.com/mrphys/Image_Reconstruction_Fractal.

*Julia Fractal Generation*

A Julia fractal is a pattern generated by repeated application of the nonlinear complex-valued function: $f(z) = z^2 + c$, where *z* is a spatial coordinate in a complex domain, and *c* is a user-defined parameter that controls the structure of the fractal. Although Julia fractals are usually generated in the 2D complex domain, simulation of real-time MRI data requires two spatial and one temporal dimension. Thus, we generated Julia fractals in quaternion space (4D complex domain) and extracted a 3D sub-volume to represent 2D+time.

A single Julia fractal was generated via the following steps:

1) A 4D grid with uniform grid spacing that reflects the dimensions of the real data $(X, Y, Z, T) \in [-1,1] \times [-1,1] \times [-0.5,0.5] \times [-0.2,0.2]$ was created, and a c value was selected (see below).
2) For each grid-point with coordinates $(x, y, z, t) \in (X, Y, Z, T)$:
   a. The quaternion $q_0$ was initialised to the coordinate vector ($q_0 = x + y\boldsymbol{i} + z\boldsymbol{j} + t\boldsymbol{k}$), where *i, j, k* are imaginary basis units of the quaternion.
   b. The recurrence relationship: $q_n = q_{n-1}^2 + c$, was then iteratively applied, starting with n=1.
   c. Iteration was terminated when $|q_n| > 4.0$ (empirically determined threshold), and the corresponding grid point was assigned the value of $n$ (i.e. the number of iterations to divergence).

*Selection of Parameter 'c' using the Mandelbrot Set*

A potentially unlimited number of distinct Julia fractals can be generated by varying c. However, most values produce blank or noise-like images. Thus, we used an empirical strategy to identify values of c which yielded structurally intricate patterns. Specifically, we observed that Julia fractals where the value of the central point had an iteration count between 10-30 produced the richest structures. Furthermore, these fractals could be efficiently identified using the Mandelbrot set, which is essentially a catalogue of Julia fractal behaviour. The recurrence for the Mandelbrot is the same as the Julia fractal, but $q_0$ is fixed for all grid-points ($q_0 = 0 + 0\boldsymbol{i} + 0\boldsymbol{j} + 0\boldsymbol{k}$), while c is set to each grid coordinate ($c = x + y\boldsymbol{i} + z\boldsymbol{j} + t\boldsymbol{k}$). Grid-points in Mandelbrot set with values between 10-30 are identified and the associated *c* values are used to generate the Julia fractals. See supplementary information video S1 for comparison of Julia fractals generated with different c parameters.

*Conversion to Complex-Valued Dynamic Images*

From each 4D fractal, we extracted the central slice in the z-axis to represent our 2D+time data and normalised between 0 and 1. To simulate complex-valued image data, two independent sine waves with different frequencies (one fixed to 0.25Hz and the other randomly sampled between 0.25Hz and 1Hz) and random phase shifts (between 0 and $2\pi$ radians) were generated. Normalised pixel intensities were then mapped to the x-axis of these two sinusoids, with the corresponding y-values defining the real and imaginary components of each pixel. Each channel was then Gaussian blurred (σ between 0.2 and 0.4) to suppress finer fractal detail and subsequently applied unsharp masking with Gaussian smoothing (σ = 0.1 and sharpening strength α = 50) to avoid overly smooth edges.

*Simulation of multi-coil MR data*

We follow the method previously applied to natural video data[3] to obtain multi-coil data consisting of the following steps:
1) A randomly scaled and rotated elliptical mask was superimposed on the image to simulate the shape of a body.
2) Background phase was simulated by randomly generating a 6x6 matrix, upsampling it to the image size and added to the image phase.
3) Synthetic coil sensitivity maps were created using 2D Gaussian functions with random centre positions, varying standard deviations in the x and y axes, random phase offset and random intensity.

4) Multi-coil data was generated by multiplying the complex value images with the synthetic coil maps and adding random background noise to each coil and time frame.
5) Fully-sampled Cartesian multi-coil k-space data was obtained by applying the Fast Fourier Transform (FFT) to the images.

Using the above steps, a dataset of 692 paired fractal image sets was generated to match the size of the available cardiac MRI training dataset[3].

*Cardiac MRI Training Data*
We used a previously described[3] dataset that included 692 electrocardiogram-triggered, Cartesian balanced steady state free precession (bSSFP) CINE multi-coil raw acquisitions obtained in seven cardiac orientations (See Supporting Information S2). The raw data were acquired with 2×undersampling, with 44 reference lines, allowing fully-sampled multi-coil Cartesian k-space data to be reconstructed using GRAPPA.

*Training the Deep Artefact Suppression Model*
Deep artefact suppression models were separately trained on fractal data and cardiac MRI data. The architecture was based on a previously described[3] 3D UNet[4] that takes multi-coil, complex, 2D+time data as input and outputs single-channel, magnitude, 2D+time data (Full architecture details in Supporting Information figure S3).

Both the cardiac MRI and fractal data were processed in a similar fashion to create paired training data that mirrored the prospective data. The fully-sampled Cartesian multi-coil k-space data was compressed to 10 coils using singular value decomposition and resampled onto 13 sorted golden angle radial spokes per time frame. This radially undersampled multi-coil k-space data was then non-uniformly fast Fourier transformed (NUFFT) to produce multi-coil, complex 2D+time image data, that was normalised so that the coil combined magnitude image data was in the range [0, 1]. This data was input into the 3D UNet with the ground truth target being the normalised coil combined (root-sum-of-squares), 2D+time, magnitude data derived from the fully-sampled k-space data.

The separate 3D UNet models were trained with a training/validation/test split of 519/69/104 for 300 epochs with the Structural Similarity Index Measure (SSIM)[5] loss and Adam[6] optimizer with learning rate set to $10^{-4}$. Training was performed on a single NVIDIA GeForce RTX 4090 and the training code and corresponding pre-trained networks are available at: https://github.com/mrphys/Image_Reconstruction_Fractal.

*Prospective Data Collection*
10 patients referred for clinical cardiac MRI were imaged on a 1.5T MRI scanner (Avanto, Siemens Medical Solutions, Erlangen, Germany). Full demographics and diagnoses are in Supporting Information Text S4. Exclusion criteria were inability to breath-hold and arrhythmia. This study was approved by the UK Health Research Authority and written informed consent was obtained in all subjects (ref.REC19/LO/1561).

A short-axis, free breathing, sorted golden radial real-time bSSFP stack (10-15 slices) was acquired (details in Supporting Information S5) and reconstructed using DL models trained with both fractal data (F-DL) and Cardiac MRI data (CMR-DL). We also reconstructed the real-time data with alternate reconstruction approaches for comparison, including: (i) Low-rank deep image prior (LR-DIP)[7]; and (ii) Conventional CS using temporal Total Variation (regularization

$5 \times 10^{-4}$). LR-DIP is a state-of-the-art unsupervised DL technique that requires no training data. We used a subspace rank of 40 and all other parameters were set to the defaults[7] with the model being trained for 2000 epochs for each image series. All real-time reconstructions were temporally interpolated to 30 frames for further analysis.

In addition, reference-standard quantification of ventricular volumes was derived from a short axis, breath-hold bSSFP CINE stack (details in Supporting Information S6).

*Prospective Data Analysis*
The four reconstructions (F-DL, CMR-DL, LR-DIP and CS) of the prospective real-time data were compared in terms of reconstruction time and image quality. They were also compared to reference-standard breath-hold CINE imaging for quantification of ventricular volumes and function.

Qualitative image evaluation: The mid-ventricular slice from all four reconstruction methods was assessed by two CMR specialists (MQ, RV). All reconstructions were presented in a single CINE clip arranged in a 4×1 side-by-side layout, with the order randomly shuffled, blinded to the scorers. The scorers ranked the images based on subjective image quality, from best (1) to worst (4) with possibility of tied scores.

Quantitative image scoring evaluation: Reconstructions of real-time data were compared using Contrast-to-Noise Ratio (CNR), edge sharpness (ES) and the temporal standard deviation (STD) of ES. The CNR was calculated as the difference in mean pixel intensity between the Left Ventricular (LV) blood-pool and myocardium, divided by the standard deviation of pixel intensities in the lung across time. The ES was measured across the boundary between LV myocardium and LV blood-pool via parametric modelling[8], and the temporal STD ES was the standard deviation of ES across time, with a lower value indicating more consistent sharpness. All measurements were repeated three times at different positions across the boundary between LV blood-pool and myocardium for each patient.

Comparison of Volumes and Function: Left ventricular and right ventricular (RV) end-diastolic volume (EDV), end-systolic volume (ESV) and ejection fraction (EF) were calculated for each reconstruction and compared to reference-standard breath-hold CINE data. For all data, preliminary segmentations (all slices and frames) were performed using a pretrained nnUNet model[9] that previously won the M&Ms challenge[10]. These were then manually corrected by a CMR expert (VM) using an open-source in-house application (Available at: https://github.com/Ti-Yao/Roundel-Biventricular) to obtain the final segmentation masks. The papillary muscles and trabeculae were included in the blood-pool.

*Statistics*
For each reconstruction type, the average qualitative rank was computed, and Friedman's test with post-hoc Nemenyi analysis was conducted. The CNR and ES measurements were subjected to Repeated Measures Analysis of Variance (RM-ANOVA) tests. The clinical metrics for each reconstruction type were compared against the corresponding metric from the reference-standard breath-hold data using Bland-Altman plots and paired t-tests with Holm correction, and two-way random-effects Intraclass Correlation Coefficient, ICC(2,1) was also computed. Statistical significance was p<0.05. Differences between the ICC of reconstruction methods were evaluated using nonparametric bootstrap resampling (5,000 iterations), and statistical significance was inferred when the 95% confidence interval of the ICC difference did not include zero.

**Results**

*Fractal Generation and Training of Deep Artefact Suppression*
Each multi-coil, complex, 2D+time fractal dataset took 11.5s to generate (~134 minutes to generate the whole fractal dataset, with 692 examples). The training time for both the F-DL and CMR-DL 3D UNets was 8hrs. Results of evaluation on simulated data are shown in Supporting Information Table S7.

*Prospective study*
Inference time: CMR-DL and F-DL had identical inference times of ~400ms per slice (all frames), resulting in a total reconstruction time of ~4-6s for all SAX slices per subject. The CS reconstruction took ~15s per slice (all frames), resulting in ~2-3 minutes for the whole SAX stack. Finally, LR-DIP reconstruction required ~30 mins per slice (all frames), leading to a total reconstruction time of ~5-8 hours.

Image Quality: Comparison of the four reconstructions (F-DL, CMR-DL, LR-DIP, CS) for the mid-ventricular short axis slice are shown in Figure 2 (with corresponding videos in Supporting Information Video S8). The F-DL and CMR-DL ranked highest in qualitative assessment with no statistical difference (p = 0.900) between them, while both were statistically better than LR-DIP and CS (p<0.001) – Table 1.

There was no significant difference in ES between F-DL, CMR-DL and CS reconstructions, although the latter two were significantly higher than LR-DIP (p<0.020). F-DL reconstructions had the lowest temporal STD ES and was significantly lower than CS (p=0.001). F-DL reconstructions had the highest CNR and was statistically significantly higher than all other reconstructions (p<0.003).

Volumes and function: The Bland-Altman plots showing agreement in clinical metrics from all reconstruction methods compared to reference-standard breath-hold data are shown in Figure 3, with bias and limits of agreement reported in Table 2. For the LV/RV volumes and EF, there were no significant biases (p>0.080) and acceptable limits of agreement for the F-DL, CMR-DL and CS reconstructions. However, the LR-DIP measurements had much wider limits of agreement and significant bias compared to reference-standard breath-hold data for LV ESV (p=0.019).

ICCs for RV/LV volumes and function were high for both F-DL and CMR-DL reconstructions compared to reference-standard breath-hold data (Table 2). However, it should be noted that LR-DIP reconstructions had significantly lower ICCs for LV EDV, LV ESV and LV EF, as well as RV EDV and RV EF.

**Discussion**
The main finding of this proof-of-concept study was that DL-based reconstructions can be trained using synthetically generated fractal data. Furthermore, these reconstructions yielded comparable image quality and clinical measures to DL-models trained on actual CMR data.

Although open-source CMR raw data repositories are becoming increasingly available (e.g. CMRXrecon[11]), we believe our fractal method has some key advantages. Firstly, we can generate almost unlimited amounts of synthetic k-space data, which may be important for more

'data hungry' models (e.g. transformers). Secondly, fractals are not limited to a single anatomy and could be used for a range of different real-time applications (e.g. speech or bowel imaging). Finally, our fractals are natively 4D and could be used to train models for 3D+t data (e.g. 3D CINE or time resolved angiography) where there is a paucity of high-quality training data. Furthermore, unlike natural videos, fractals may elicit less clinical scepticism because they are abstract and mathematically generated and free from incongruous real-world content (e.g., moving vehicles).

In our study, prospective real-time MRI data reconstructed using F-DL and CMR-DL demonstrated comparable image quality. This is consistent with our work on natural videos[3], and is most likely due to deep artefact suppression primarily 'learning' local image features that are similar in both cardiac MRI and fractal data. We also showed that LV and RV volumes and function derived from F-DL and CMR-DL reconstructions of real-time data had comparable agreement with reference-standard breath-hold CINE imaging. This is vital as demonstration of clinical relevance is a pre-requisite of future wider use.

We also compared supervised DL methods to competing approaches including LR-DIP and conventional CS. LR-DIP is of great interest as it promises some of the benefits of DL without the need for training data. However, in our study, both supervised methods (F-DL and CMR-DL) slightly outperformed both LR-DIP and CS. More importantly, the reconstruction times for CS and LR-DIP were both significantly longer than deep artefact suppression, with LR-DIP taking hours to reconstruct a SAX stack. This substantially hampers clinical use and strengthens the case for supervised DL, and consequently for the use of synthetic training data. It should be noted that newer iterative self-supervised methods do have lower reconstruction times[12], but they could also potentially benefit from pretraining on synthetic data.

This study has several limitations, reflecting its proof-of-concept nature. Firstly, only one sampling pattern and one DL method were tested, and further work is required to demonstrate that fractal-generated training data generalises across different trajectories and reconstruction architectures. However, our previous work on natural videos does suggest that synthetic data is highly generalisable. Secondly, other studies have demonstrated different methods for generating synthetic phase from magnitude-only images[13], [14]. Improving this component of the framework may benefit some applications and needs investigation. Finally, the patient cohort was small, and larger more diverse population are required for comprehensive clinical validation.

*Conclusion*
 We developed and demonstrated using self-generated fractal data to train DL models for dynamic MRI reconstruction. We belive this method  has strong potential for training highly generalisable dynamic MRI reconstruction models in the future.

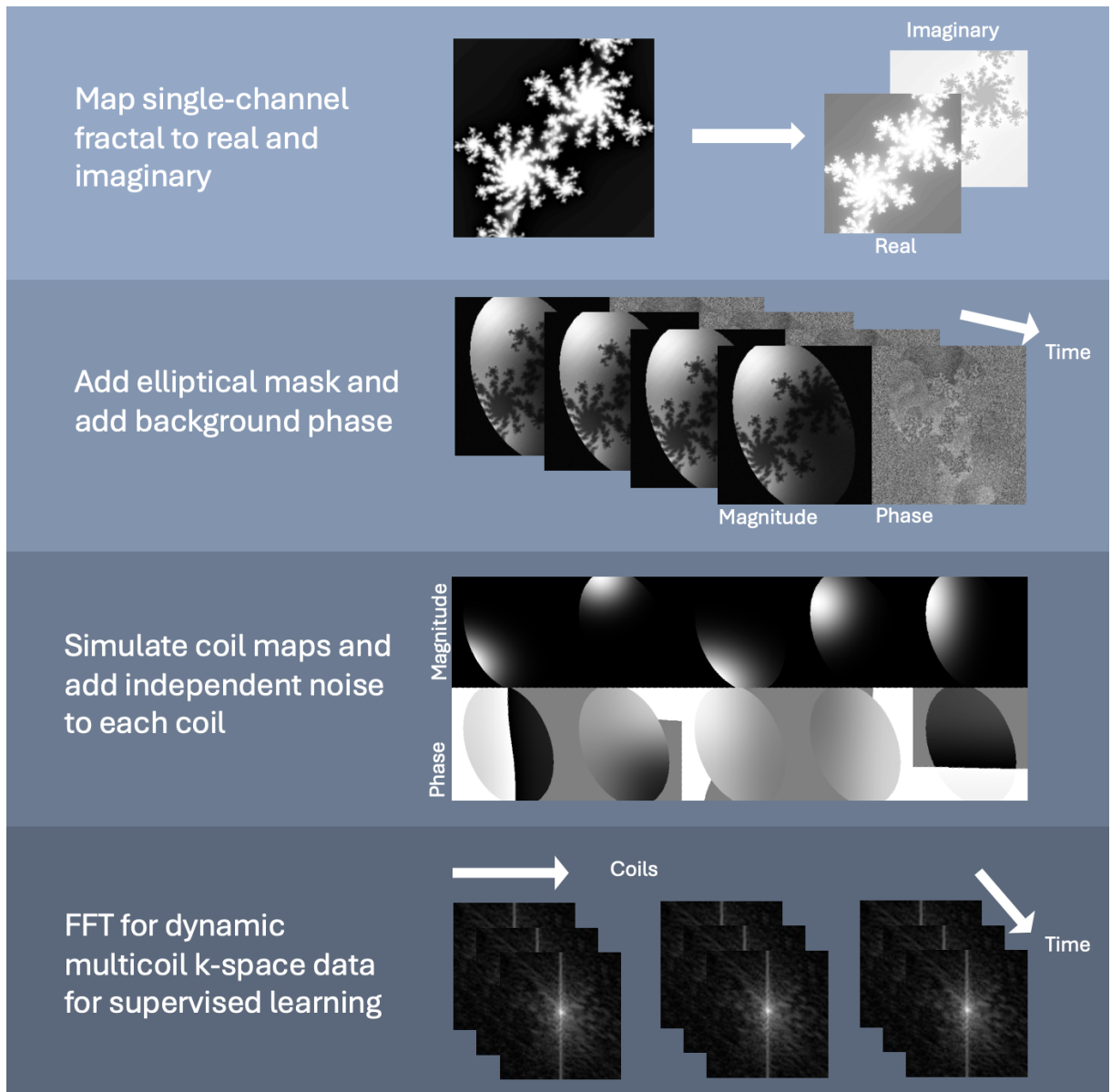

*Figure 1: Process for creating multi-coil complex data from fractals: i) Map single-channel fractals to 2 channels (real and imaginary) using two sine waves; ii) Simulate shape of a body using elliptical masking and add low-frequency background phase; iii) Simulate coils using Gaussian intensity distributions and add independent noise to each coil; and iv) Apply a fast Fourier Transformation to acquire data in k-space.*

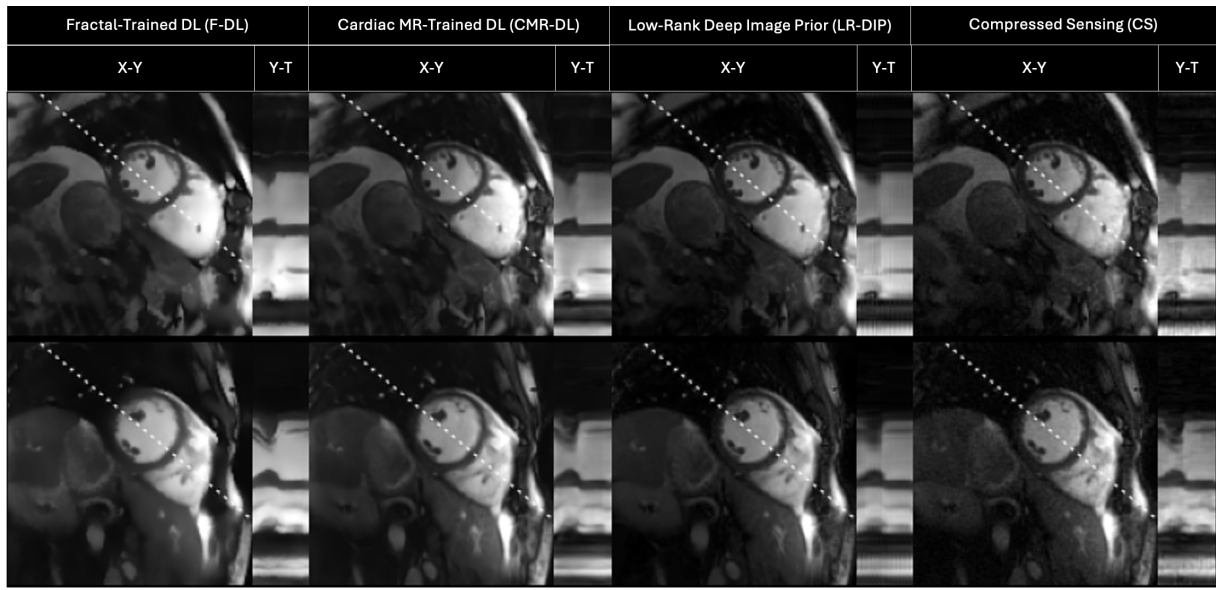

Figure 2: Comparison of reconstructions of real-time MRI data using the four methods (left to right: F-DL, CMR-DL, LR-DIP, CS) with X-Y plots of the mid-ventricular short-axis slice at end-diastole and Y-T plots of a line across the ventricular septum through time.

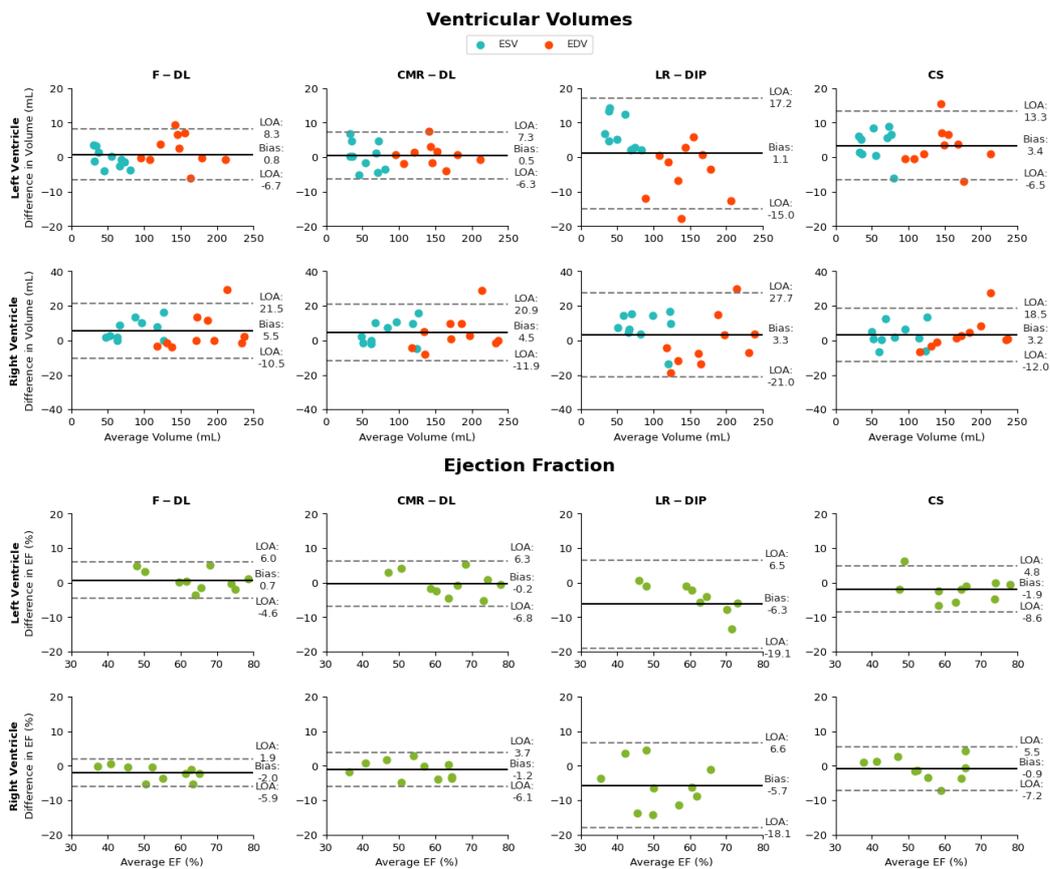

Figure 3: Bland-Altman plots of real-time data with all four reconstructions (F-DL, CMR-DL, LR-DIP, CS) compared to reference-standard breath-hold data for measurements of end-systolic volume (ESV), end-diastolic volume (EDV) and ejection fraction (EF) for the left ventricle (LV) and right ventricle (RV).

Table 1: Image quality comparison of real-time data from all four reconstruction methods (F-DL, CMR-DL, LR-DIP and CS) and reference-standard breath-hold (BH) images, with Edge Sharpness (ES) in mm$^{-1}$, Temporal STD ES, Contrast-to-Noise Ratio (CNR) and subjective image quality ranking (1=best, 4=worst). The superscripts †, ✦, ▲, °, * indicate statistically significant inferiority to F-DL, CMR-DL, LR-DIP, CS and BH respectively.

| Image Type | Edge Sharpness (ES) | Temporal STD ES | Contrast-to-Noise Ratio (CNR) | Subjective Image Quality Ranking |
|---|---|---|---|---|
| F-DL | 0.201 ± 0.073 | 0.0511 ± 0.0336 | 127.3 ± 103.7 | 1.6 ± 0.6 |
| CMR-DL | 0.211 ± 0.069 | 0.0634 ± 0.0510 | 72.5 ± 68.3† | 1.6 ± 0.7 |
| LR-DIP | 0.188 ± 0.079✦° | 0.0619 ± 0.0548 | 42.6 ± 26.2†✦ | 3.6 ± 0.6†✦ |
| CS | 0.218 ± 0.083 | 0.0712 ± 0.0479† | 25.4 ± 8.3†✦▲* | 3.0 ± 0.9†✦ |

Table 2: Clinical metric comparison of measurements from real-time data with all four reconstructions (F-DL, CMR-DL, LR-DIP and CS) with Bias, Limits of Agreement (LOA) and Intraclass Correlation Coefficient (ICC) between reconstructions and reference-standard breath-hold data. Measurements included End-Systolic Volume (ESV), End-Diastolic Volume (EDV) and Ejection fraction (EF) of the Left Ventricle (LV) and Right Ventricle (RV). ▲ indicates statistically significant bias compared to reference-standard breath-hold data, and statistically significant difference between reconstructions are indicated ✷ for difference with F-DL, † for CMR-DL and ✦ for CS.

| Metric | Reconstruction | Left Ventricle | | | Right Ventricle | | |
|---|---|---|---|---|---|---|---|
| | | Bias | LOA | ICC | Bias | LOA | ICC |
| End-Systolic Volume | F-DL | -0.49 mL | 4.43 to -5.4 mL | 0.99 (0.97 to 1) | 6.32 mL | 17.38 to -4.75 mL | 0.96 (0.62 to 0.99) |
| | CMR-DL | 0.3 mL | 7.93 to -7.33 mL | 0.98 (0.92 to 0.99) | 4.74 mL | 17.5 to -8.03 mL | 0.97 (0.83 to 0.99) |
| | LR-DIP | 6.61 mL | 15.71 to -2.5 mL | 0.91 (0.15 to 0.98)†✷ | 7.84 mL ▲ | 24.5 to -8.82 mL | 0.92 (0.54 to 0.98)†✦ |
| | CS | 3.78 mL | 12.27 to -4.72 mL | 0.96 (0.74 to 0.99)✷ | 2.92 mL | 15.43 to -9.6 mL | 0.97 (0.9 to 0.99) |
| End-Diastolic Volume | F-DL | 2.15 mL | 10.78 to -6.49 mL | 0.99 (0.96 to 1) | 4.67 mL | 24.23 to -14.88 mL | 0.97 (0.87 to 0.99) |
| | CMR-DL | 0.68 mL | 6.51 to -5.15 mL | 0.99 (0.98 to 1) | 4.28 mL | 23.57 to -15.02 mL | 0.97 (0.88 to 0.99) |
| | LR-DIP | -4.43 mL | 9.82 to -18.67 mL | 0.97 (0.87 to 0.99)† | -1.19 mL | 26.25 to -28.64 mL | 0.95 (0.81 to 0.99) |
| | CS | 3.05 mL | 14.14 to -8.03 mL | 0.98 (0.93 to 1) | 3.53 mL | 21.1 to -14.05 mL | 0.97 (0.91 to 0.99) |
| Ejection Fraction | F-DL | 0.69 % | 6 to -4.62 % | 0.96 (0.86 to 0.99) | -2.02 % | 1.9 to -5.95 % | 0.96 (0.67 to 0.99) |
| | CMR-DL | -0.23 % | 6.35 to -6.8 % | 0.95 (0.8 to 0.99) | -1.18 % | 3.72 to -6.08 % | 0.96 (0.86 to 0.99) |
| | LR-DIP | -6.26 % | 6.54 to -19.06 % | 0.65 (0.01 to 0.9)†✷ | -5.75 % | 6.57 to -18.07 % | 0.69 (0.06 to 0.92)†✷✦ |
| | CS | -1.91 % | 4.77 to -8.6 % | 0.93 (0.74 to 0.98) | -0.85 % | 5.48 to -7.18 % | 0.95 (0.81 to 0.99) |